\documentclass[11pt]{article}
\usepackage{acl2016}
\usepackage{times}
\usepackage{latexsym}

\usepackage{bm}
\usepackage{amsmath, amssymb}
\usepackage[pdftex]{graphicx}
\usepackage{url}
\usepackage{booktabs}
\usepackage{multirow}
\usepackage{CJKutf8}

\aclfinalcopy 

\title{Tree-to-Sequence Attentional Neural Machine Translation}

\author{
 Akiko Eriguchi, Kazuma Hashimoto, and Yoshimasa Tsuruoka\\
 The University of Tokyo, 3-7-1 Hongo, Bunkyo-ku, Tokyo, Japan\\
 {\tt \{eriguchi, hassy, tsuruoka\}@logos.t.u-tokyo.ac.jp}\\
}
\date{}
\begin{document}
\maketitle
\begin{abstract}
Most of the existing Neural Machine Translation (NMT) models focus on the conversion of sequential data and do not directly use syntactic information. 
We propose a novel end-to-end syntactic NMT model, extending a sequence-to-sequence model with the source-side phrase structure. 
Our model has an attention mechanism that enables the decoder to generate a translated word while softly aligning it with phrases as well as words of the source sentence. 
Experimental results on the WAT'15 English-to-Japanese dataset demonstrate that our proposed model considerably outperforms sequence-to-sequence attentional NMT models and compares favorably with the state-of-the-art tree-to-string SMT system.
\end{abstract}

\section{Introduction} \label{Introduction}
Machine Translation (MT) has traditionally been one of the most complex language processing problems, but recent advances of Neural Machine Translation (NMT) make it possible to perform translation using a simple end-to-end architecture. 
In the Encoder-Decoder model~\cite{cho-EtAl:2014:EMNLP2014,NIPS2014_5346}, a Recurrent Neural Network (RNN) called the {\it encoder} reads the whole sequence of source words to produce a fixed-length vector, and then another RNN called the {\it decoder} generates the target words from the vector. 
The Encoder-Decoder model has been extended with an {\it attention} mechanism~\cite{DBLP:journals/corr/BahdanauCB14,luong-pham-manning:2015:EMNLP}, which allows the model to jointly learn the soft alignment between the source language and the target language.
NMT models have achieved state-of-the-art results in English-to-French and English-to-German translation tasks~\cite{luong-EtAl:2015:ACL-IJCNLP,luong-pham-manning:2015:EMNLP}. 
However, it is yet to be seen whether NMT is competitive with traditional Statistical Machine Translation (SMT) approaches in translation tasks for structurally distant language pairs such as English-to-Japanese.

\begin{figure}[t]
  \begin{center}
    \includegraphics[clip,width=5.0cm]{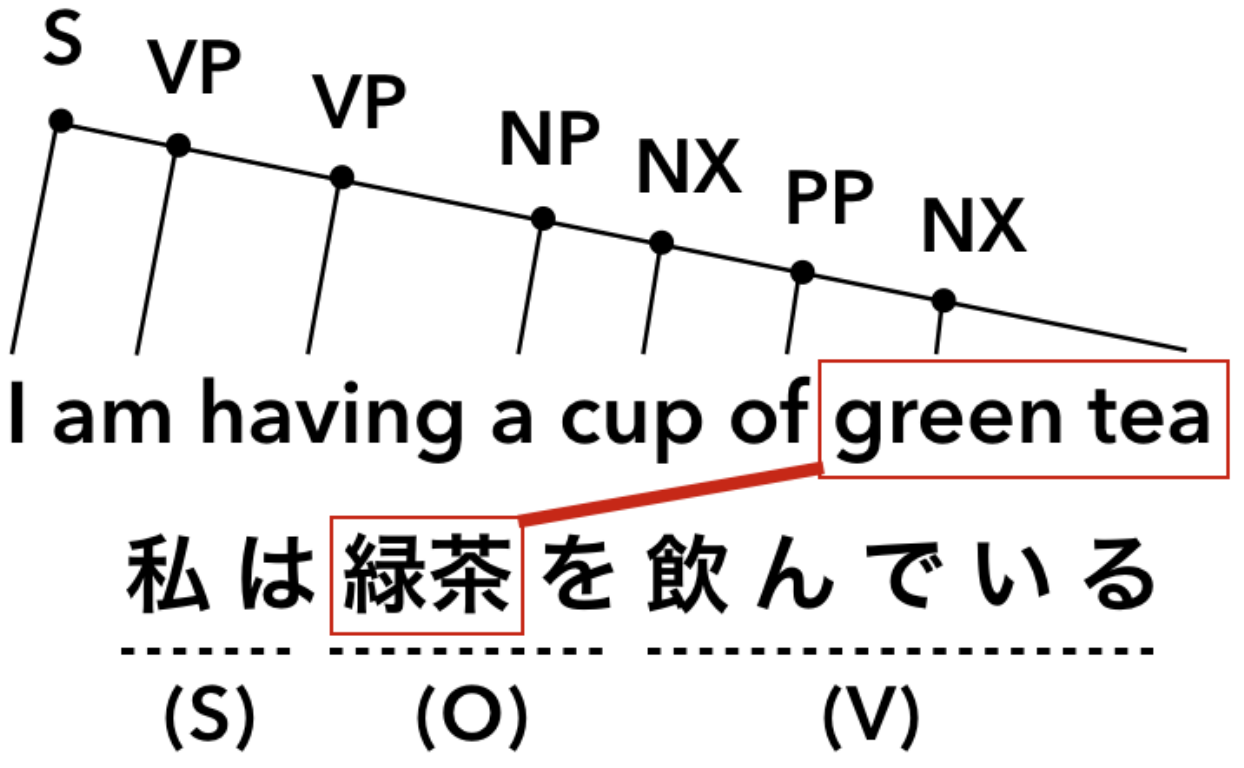}
    \caption{Alignment between an English phrase and a Japanese word.}
   \label{fig: align_phrase}
  \end{center}
\end{figure}

Figure~\ref{fig: align_phrase} shows a pair of parallel sentences in English and Japanese. 
English and Japanese are linguistically distant  in many respects; they have different syntactic constructions, and words and phrases are defined in different lexical units. 
In this example, the Japanese word ``\begin{CJK}{UTF8}{min}緑茶\end{CJK}" is aligned with the English words \lq\lq green" and \lq\lq tea", and the English word sequence \lq\lq a cup of" is aligned with a special symbol \lq\lq {\it null}", which is not explicitly translated into any Japanese words. 
One way to solve this mismatch problem is to consider the phrase structure of the English sentence and align the phrase ``a cup of green tea" with ``\begin{CJK}{UTF8}{min}緑茶\end{CJK}". 
In SMT, it is known that incorporating syntactic constituents of the source language into the models improves word alignment~\cite{Yamada:2001:SST:1073012.1073079} and translation accuracy~\cite{liu-liu-lin:2006:COLACL,neubig-duh:2014:P14-2}.
However, the existing NMT models do not allow us to perform this kind of alignment.

In this paper, we propose a novel attentional NMT model to take advantage of syntactic information.
Following the phrase structure of a source sentence, we encode the sentence recursively in a bottom-up fashion to produce a vector representation of the sentence and decode it while aligning the input phrases and words with the output.
Our experimental results on the WAT'15 English-to-Japanese translation task show that our proposed model achieves state-of-the-art translation accuracy.

\section{Neural Machine Translation} 
\subsection{Encoder-Decoder Model} \label{subsection: encoder-decoder}
NMT is an end-to-end approach to data-driven machine translation~\cite{kalchbrenner-blunsom:2013:EMNLP,NIPS2014_5346,DBLP:journals/corr/BahdanauCB14}. 
In other words, the NMT models directly estimate the conditional probability $p({\bm y}| {\bm x})$ given a large collection of source and target sentence pairs $({\bm x}, {\bm y})$.
An NMT model consists of an encoder process and a decoder process, and hence they are often called {\it Encoder-Decoder} models.
In the Encoder-Decoder models, a sentence is treated as a sequence of words. 
In the encoder process, the encoder embeds each of the source words ${\bm x} = (x_{1}, x_{2}, \cdots, x_{n})$ into a $d$-dimensional vector space. 
The decoder then outputs a word sequence ${\bm y} = (y_{1}, y_{2}, \cdots, y_{m})$ in the target language given the information on the source sentence provided by the encoder.
Here, $n$ and $m$ are the lengths of the source and target sentences, respectively.
RNNs allow one to effectively embed sequential data into the vector space.

In the RNN encoder, the $i$-th hidden unit ${\bm h}_{i} \in \mathbb{R}^{d\times 1}$ is calculated given the $i$-th input $x_{i}$ and the previous hidden unit ${\bm h}_{i-1} \in \mathbb{R}^{d\times 1}$,
 \begin{eqnarray}
  {\bm h}_{i} & = &
	f_{enc}(x_{i}, {\bm h}_{i-1}),				\label{eq: encoder_definition}
\end{eqnarray}
where $f_{enc}$ is a non-linear function, and the initial hidden unit ${\bm h}_{0}$ is usually set to zeros.
The encoding function $f_{enc}$ is recursively applied until the $n$-th hidden unit ${\bm h}_{n}$ is obtained.
The RNN Encoder-Decoder models assume that ${\bm h}_{n}$ represents a vector of the meaning of the input sequence up to the $n$-th word.

After encoding the whole input sentence into the vector space, we decode it in a similar way. 
The initial decoder unit ${\bm s}_{1}$ is initialized with the input sentence vector (${\bm s}_{1} = {\bm h}_{n})$.
Given the previous target word and the $j$-th hidden unit of the decoder, the conditional probability that the $j$-th target word is generated is calculated as follows:
 \begin{eqnarray}
  p(y_{j} | {\bm y}_{<j}, {\bm x}) & = &
	g({\bm s}_{j}),				\label{eq: output_layer}
\end{eqnarray}
where $g$ is a non-linear function.
The $j$-th hidden unit of the decoder is calculated by using another non-linear function $f_{dec}$ as follows: 
 \begin{eqnarray}
 {\bm s}_{j} & = &
	f_{dec}(y_{j-1}, {\bm s}_{j-1}).					\label{eq: decoder_calculation}	
\end{eqnarray}

We employ Long Short-Term Memory (LSTM) units~\cite{Hochreiter:1997:LSM:1246443.1246450,journals/neco/GersSC00} in place of vanilla RNN units.
The $t$-th LSTM unit consists of several {\it gates} and two different types of states: a hidden unit ${\bm h}_{t} \in \mathbb{R}^{d\times 1}$ and a memory cell ${\bm c}_{t} \in \mathbb{R}^{d\times 1}$, 
\begin{eqnarray}
{\bm i}_{t} & = & 
	\sigma ( {\bm W}^{(i)} {\bm x}_{t} + {\bm U}^{(i)} {\bm h}_{t-1} + {\bm b}^{(i)}), 		\nonumber	\\ 
{\bm f}_{t} & = & 
	\sigma ( {\bm W}^{(f)} {\bm x}_{t} + {\bm U}^{(f)} {\bm h}_{t-1} + {\bm b}^{(f)}),	 	\nonumber	\\ 
{\bm o}_{t} & = &
	\sigma ( {\bm W}^{(o)} {\bm x}_{t} + {\bm U}^{(o)} {\bm h}_{t-1} + {\bm b}^{(o)}), 	\nonumber	\\
\tilde{{\bm c}}_{t} & = &
	\tanh ( {\bm W}^{(\tilde{c})} {\bm x}_{t} + {\bm U}^{(\tilde{c})} {\bm h}_{t-1} + {\bm b}^{(\tilde{c})}), 	\nonumber	\\
{\bm c}_{t} & = & 
	{\bm i}_{t} \odot \tilde{{\bm c}}_{t} + {\bm f}_{t} \odot {\bm c}_{t-1}, 				\nonumber	\\
{\bm h}_{t} & = &
	{\bm o}_{t} \odot \tanh ({\bm c}_{t}),						\label{eq: LSTM}
\end{eqnarray}
where each of ${\bm i}_{t}$, ${\bm f}_{t}$, ${\bm o}_{t}$ and $\tilde{{\bm c}}_{t} \in \mathbb{R}^{d\times 1}$ denotes an input gate, a forget gate, an output gate, and a state for updating the memory cell, respectively.
${\bm W}^{(\cdot)} \in \mathbb{R}^{d\times d}$ and ${\bm U}^{(\cdot)} \in \mathbb{R}^{d\times d}$ are weight matrices, ${\bm b}^{(\cdot)}\in \mathbb{R}^{d\times1}$ is a bias vector, and $\bm{x}_t\in\mathbb{R}^{d\times 1}$ is the word embedding of the $t$-th input word.
$\sigma (\cdot)$ is the logistic function, and the operator $\odot$ denotes element-wise multiplication between vectors.

\subsection{Attentional Encoder-Decoder Model}
The NMT models with an attention mechanism~\cite{DBLP:journals/corr/BahdanauCB14,luong-pham-manning:2015:EMNLP}
have been proposed to softly align each decoder state with the encoder states. 
The attention mechanism allows the NMT models to explicitly quantify how much each encoder state contributes to the word prediction at each time step.

\begin{figure}[t]
  \begin{center}
  \includegraphics[clip,width=5.0cm]{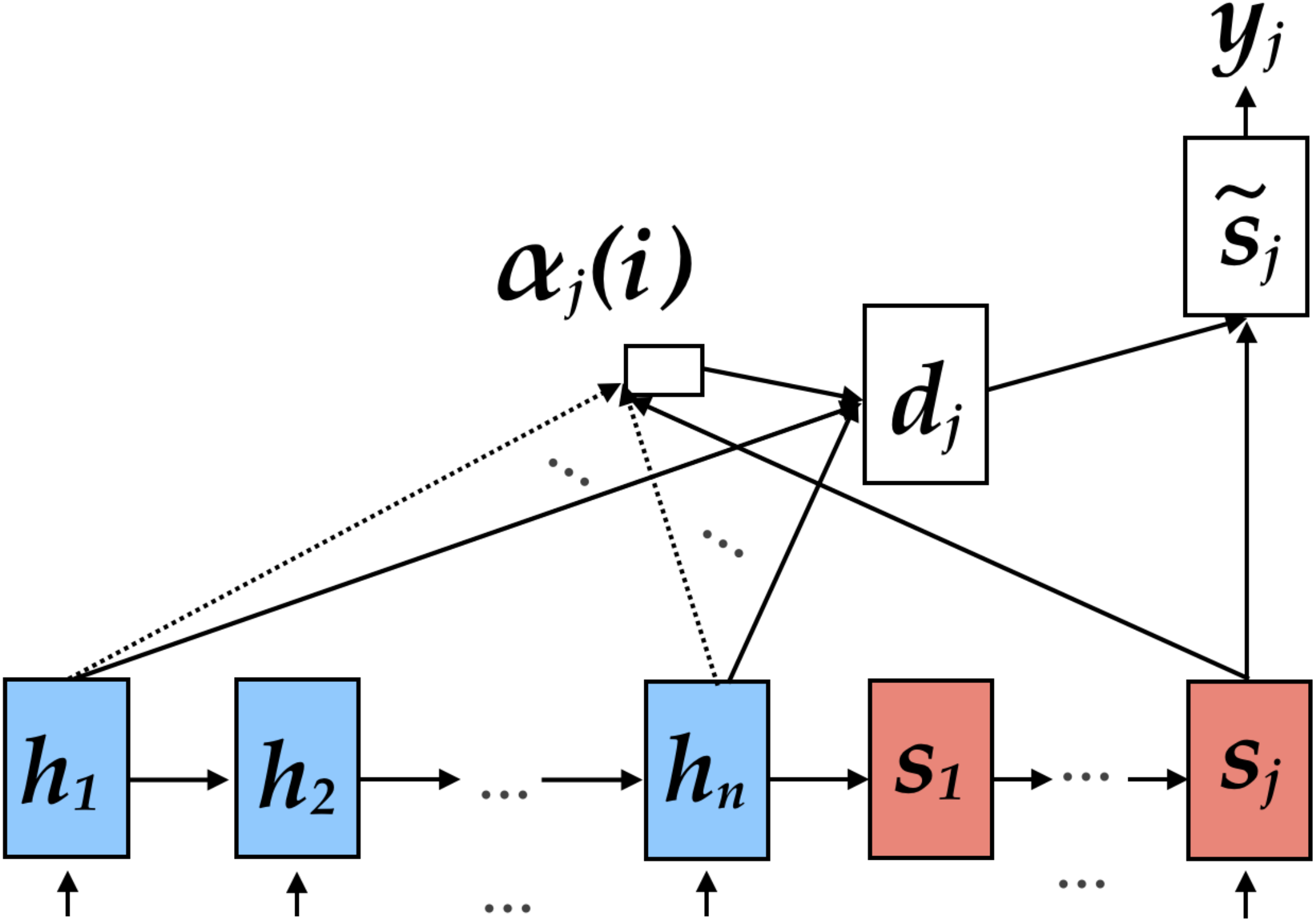}
    \caption{Attentional Encoder-Decoder model.}
   \label{fig: attention_model}
  \end{center}
\end{figure}

In the attentional NMT model in \newcite{luong-pham-manning:2015:EMNLP}, 
at the $j$-th step of the decoder process, the attention score ${\bm \alpha}_{j}(i)$ between the $i$-th source hidden unit ${\bm h}_{i}$ and the $j$-th target hidden unit ${\bm s}_{j}$ is calculated as follows: 
\begin{eqnarray} 
{\bm \alpha}_{j}(i) 
	& = & \frac
		{\exp ({\bm h}_{i} \cdot {\bm s}_{j})}
		{\sum_{k=1}^{n} \exp ({\bm h}_{k} \cdot {\bm s}_{j})},
	\label{eq: attentionWeight}
\end{eqnarray}
where ${\bm h}_{i} \cdot {\bm s}_{j}$ is the inner product of ${\bm h}_{i}$ and ${\bm s}_{j}$, which is used to directly calculate the similarity score between ${\bm h}_{i}$ and ${\bm s}_{j}$.
The $j$-th context vector ${\bm d}_{j}$ is calculated as the summation vector weighted by ${\bm \alpha}_{j}(i)$:
\begin{eqnarray} 
{\bm d}_{j} 
	& = & \sum_{i=1}^{n} {\bm \alpha}_{j}(i) {\bm h}_{i}.
\label{eq: context_calc}	
\end{eqnarray}
To incorporate the attention mechanism into the decoding process, the context vector is used for the the $j$-th word prediction by putting an additional hidden layer $\tilde{{\bm s}}_{j}$:
\begin{eqnarray} 
\tilde{{\bm s}}_{j} 
	= \tanh ({\bm W}_{d} [{\bm s}_{j}; ~{\bm d}_{j}] + {\bm b}_{d}),
			\label{eq: final_Decoder_Luong}	
\end{eqnarray}
where $[{\bm s}_{j}; {\bm d}_{j}] \in \mathbb{R}^{2d \times 1}$ is the concatenation of ${\bm s}_{j}$ and ${\bm d}_{j}$, and ${\bm W}_{d} \in \mathbb{R}^{d\times 2d}$ and ${\bm b_{d}} \in \mathbb{R}^{d\times 1}$ are a weight matrix and a bias vector, respectively. 
The model predicts the $j$-th word by using the softmax function:
\begin{eqnarray}
p(y_{j} | {\bm y}_{<j}, {\bm x}) = \mbox{softmax} ({\bm W}_{s} \tilde{{\bm s}}_{j} + {\bm b}_{s}), 
\label{eq: decoder_softmax}
\end{eqnarray}
where ${\bm W}_{s} \in \mathbb{R}^{|V|\times d}$ and ${\bm b_{s}}\in \mathbb{R}^{|V|\times 1}$ are a weight matrix and a bias vector, respectively.
$|V|$ stands for the size of the vocabulary of the target language.
Figure~\ref{fig: attention_model} shows an example of the NMT model with the attention mechanism.

\subsection{Objective Function of NMT Models}
The objective function to train the NMT models is the sum of the log-likelihoods of the translation pairs in the training data: 
\begin{eqnarray} 
J({\bm \theta})
	& = & \frac{1}{|\mathcal{D} |} \sum_{({\bm x}, {\bm y}) \in \mathcal{D}} \log p({\bm y} | {\bm x}), 		\label{eq: objective_func}	
\end{eqnarray}
where $\mathcal{D}$ denotes a set of parallel sentence pairs.
The model parameters ${\bm \theta}$ are learned through Stochastic Gradient Descent (SGD).

\section{Attentional Tree-to-Sequence Model}
\subsection{Tree-based Encoder + Sequential Encoder}
\begin{figure}[t]
  \begin{center}
  \includegraphics[clip,width=5.0cm]{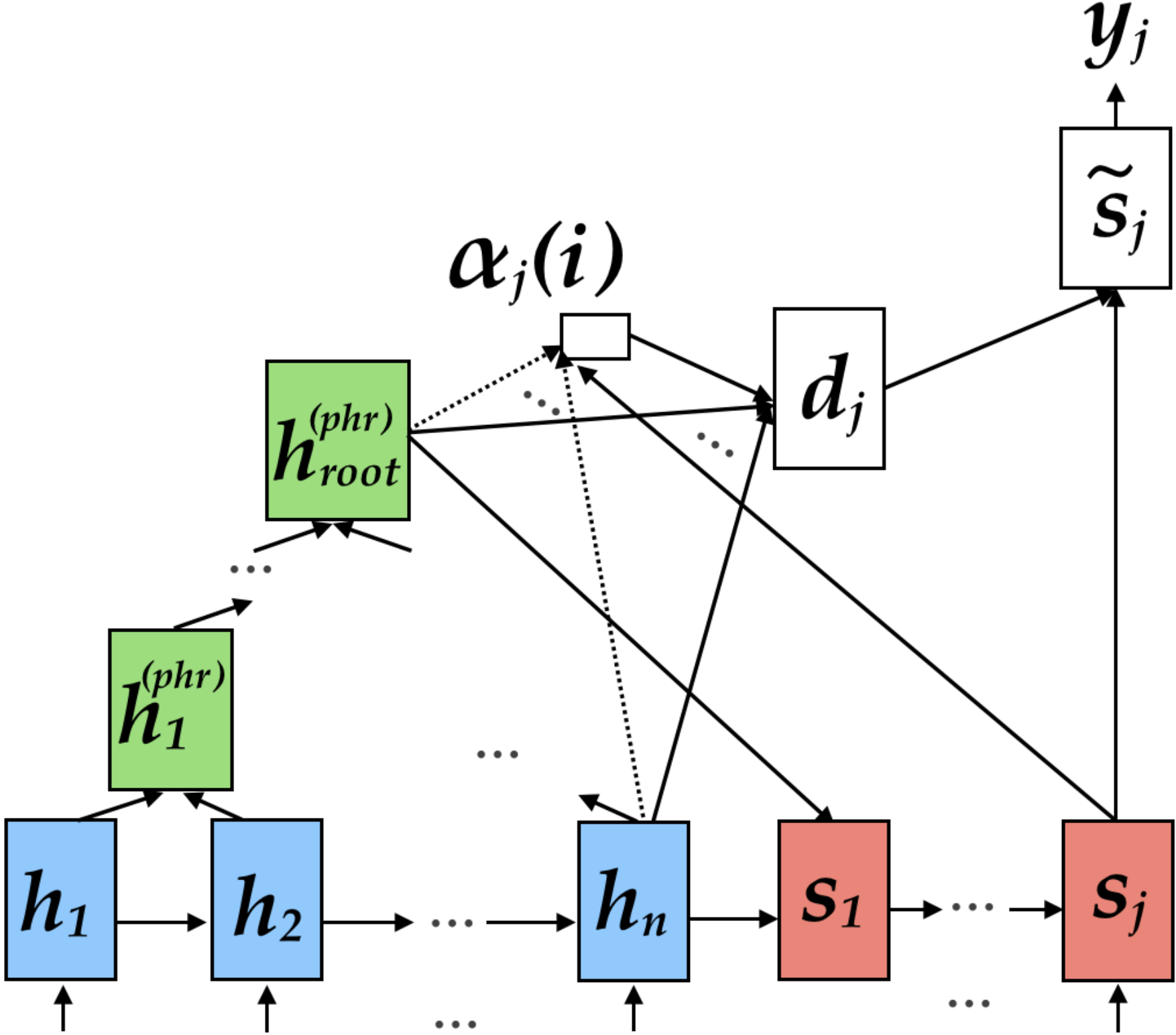}
    \caption{Proposed model: Tree-to-sequence attentional NMT model.}
   \label{fig: proposed}
     \end{center}
\end{figure}

The exsiting NMT models treat a sentence as a sequence of words and neglect the structure of a sentence inherent in language.
We propose a novel tree-based encoder in order to explicitly take the syntactic structure into consideration in the NMT model. 
We focus on the phrase structure of a sentence and construct a sentence vector from phrase vectors in a bottom-up fashion.
The sentence vector in the tree-based encoder is therefore composed of the structural information rather than the sequential data. 
Figure~\ref{fig: proposed} shows our proposed model, which we call a {\it tree-to-sequence attentional NMT model}.

In Head-driven Phrase Structure Grammar (HPSG)~\cite{sag2003syntactic}, a sentence is composed of multiple phrase units and represented as a binary tree as shown in Figure~\ref{fig: align_phrase}.
Following the structure of the sentence, we construct a tree-based encoder on top of the standard sequential encoder. 
The $k$-th parent hidden unit ${\bm h}^{(phr)}_{k}$ for the $k$-th phrase is calculated using the left and right child hidden units ${\bm h}^{l}_{k}$ and ${\bm h}^{r}_{k}$ as follows:
 \begin{eqnarray}
  {\bm h}^{(phr)}_{k} & = &
	f_{tree}({\bm h}^{l}_{k}, {\bm h}^{r}_{k}),				\label{eq: tree_encoder_definition}
\end{eqnarray}
where $f_{tree}$ is a non-linear function. 

We construct a tree-based encoder with LSTM units, where each node in the binary tree is represented with an LSTM unit.
When initializing the leaf units of the tree-based encoder,
we employ the sequential LSTM units described in Section~\ref{subsection: encoder-decoder}. 
Each non-leaf node is also represented with an LSTM unit, and we employ Tree-LSTM~\cite{tai-socher-manning:2015:ACL-IJCNLP} to calculate the LSTM unit of the parent node which has two child LSTM units.
The hidden unit ${\bm h}^{(phr)}_{k} \in \mathbb{R}^{d\times 1}$ and the memory cell ${\bm c}^{(phr)}_{k} \in \mathbb{R}^{d\times 1}$ for the $k$-th parent node are calculated  as follows:
\begin{eqnarray} 
{\bm i}_{k} & = & 
	\sigma ({\bm U}_{l}^{(i)} {\bm h}^{l}_{k} + {\bm U}_{r}^{(i)} {\bm h}^{r}_{k} + {\bm b}^{(i)}),\nonumber \\
{\bm f}^{l}_{k} & = & 
	\sigma ({\bm U}_{l}^{(f_{l})} {\bm h}^{l}_{k} + {\bm U}_{r}^{(f_{l)}} {\bm h}^{r}_{k} + {\bm b}^{(f_{l})}), 	\nonumber	\\
{\bm f}^{r}_{k} & = & 
	\sigma ({\bm U}_{l}^{(f_{r})} {\bm h}^{l}_{k} + {\bm U}_{r}^{(f_{r})} {\bm h}^{r}_{k} + {\bm b}^{(f_{r})}), 	\nonumber\\
{\bm o}_{k} & = &
	\sigma ({\bm U}_{l}^{(o)} {\bm h}^{l}_{k} + {\bm U}_{r}^{(o)} {\bm h}^{r}_{k} + {\bm b}^{(o)}), 	\nonumber\\
\tilde{{\bm c}}_{k} & = &
	\tanh ({\bm U}_{l}^{(\tilde{c})} {\bm h}^{l}_{k} + {\bm U}_{r}^{(\tilde{c})} {\bm h}^{r}_{k} + {\bm b}^{(\tilde{c})}), 	\nonumber \\
{\bm c}^{(phr)}_{k} & = & 
	{\bm i}_{k} \odot \tilde{{\bm c}}_{k} + {\bm f}^{l}_{k} \odot {\bm c}^{l}_{k} + {\bm f}^{r}_{k} \odot {\bm c}^{r}_{k}, 	\nonumber	\\
{\bm h}^{(phr)}_{k} & = &
	{\bm o}_{k} \odot \tanh ({\bm c}^{(phr)}_{k}),				 						\label{eq: TreeLSTM}
\end{eqnarray}
where ${\bm i}_{k}$, ${\bm f}^{l}_{k}$, ${\bm f}^{r}_{k}$, ${\bm o}_{j}$, $\tilde{{\bm c}}_{j} \in \mathbb{R}^{d\times 1}$ 
are an input gate, the forget gates for left and right child units, 
an output gate, and a state for updating the memory cell, respectively.
${\bm c}^{l}_{k}$ and ${\bm c}^{r}_{k}$ are the memory cells for the left and right child units, respectively.
${\bm U}^{(\cdot)} \in \mathbb{R}^{d\times d}$ denotes a weight matrix, and ${\bm b}^{(\cdot)}\in \mathbb{R}^{d\times1}$ represents a bias vector.

Our proposed tree-based encoder is a natural extension of the conventional sequential encoder, since Tree-LSTM is a generalization of chain-structured LSTM~\cite{tai-socher-manning:2015:ACL-IJCNLP}.
Our encoder differs from the original Tree-LSTM in the calculation of the LSTM units for the leaf nodes.
The motivation is to construct the phrase nodes in a context-sensitive way, which, for example, allows the model to compute different representations for multiple occurrences of the same word in a sentence because the sequential LSTMs are calculated in the context of the previous units. 
This ability contrasts with the original Tree-LSTM, in which the leaves are composed only of the word embeddings without any contextual information.

\subsection{Initial Decoder Setting}
We now have two different sentence vectors: one is from the sequence encoder and the other from the tree-based encoder. 
As shown in Figure~\ref{fig: proposed}, we provide another Tree-LSTM unit which has the final sequential encoder unit (${\bm h}_{n}$) and the tree-based encoder unit (${\bm h}^{(phr)}_{root}$) as two child units and set it as the initial decoder ${\bm s}_{1}$ as follows:
\begin{eqnarray}
    {\bm s}_{1} 
	 = g_{tree}({\bm h}_{n}, {\bm h}^{(phr)}_{root}), 
\end{eqnarray}
where $g_{tree}$ is the same function as $f_{tree}$ with another set of Tree-LSTM parameters.
This initialization allows the decoder to capture information from both the sequential data and phrase structures.
\newcite{corr/abs/1601.00710} proposed a similar method using a Tree-LSTM for initializing the decoder, with which they translate multiple source languages to one target language.
When the syntactic parser fails to output a parse tree for a sentence, we encode the sentence with the sequential encoder by setting ${\bm h}^{(phr)}_{root} = {\bm 0}$.
Our proposed tree-based encoder therefore works with any sentences.

\subsection{Attention Mechanism in Our Model}
We adopt the attention mechanism into our tree-to-sequence model in a novel way.
Our model gives attention not only to sequential hidden units but also to phrase hidden units.
This attention mechanism tells us which words or phrases in the source sentence are important when the model decodes a target word.
The $j$-th context vector ${\bm d}_{j}$ is composed of the sequential and phrase vectors weighted by the attention score ${\bm \alpha}_{j}(i)$:
\begin{eqnarray}
{\bm d}_{j} 
	 =  \sum_{i=1}^{n} {\bm \alpha}_{j}(i) {\bm h}_{i} 
                + \sum_{i=n+1}^{2n-1} {\bm \alpha}_{j}(i) {\bm h}^{(phr)}_{i}.
                \label{eq: proposed_context}
\end{eqnarray}
Note that a binary tree has $n-1$ phrase nodes if the tree has $n$ leaves. 
We set a final decoder $\tilde{{\bm s}}_{j}$ in the same way as Equation~(\ref{eq: final_Decoder_Luong}).

In addition, we adopt the {\it input-feeding} method~\cite{luong-pham-manning:2015:EMNLP} in our model, which is a method for feeding  $\tilde{{\bm s}}_{j-1}$, the previous unit to predict the word $y_{j-1}$, into the current target hidden unit ${\bm s}_{j}$,
\begin{eqnarray}
{\bm s}_{j} = 
	f_{dec} (y_{j-1}, [{\bm s}_{j-1}; ~\tilde{{\bm s}}_{j-1}]),
			\label{eq: input-feeding}
\end{eqnarray}
where $[{\bm s}_{j-1}; ~\tilde{{\bm s}}_{j-1}]$ is the concatenation of ${\bm s}_{j-1}$ and $\tilde{{\bm s}}_{j-1}$.
The input-feeding approach contributes to the enrichment in the calculation of the decoder, because $\tilde{{\bm s}}_{j-1}$ is an informative unit which can be used to predict the output word as well as to be compacted with attentional context vectors.
\newcite{luong-pham-manning:2015:EMNLP} showed that the input-feeding approach improves BLEU scores. 
We also observed the same improvement in our preliminary experiments.

\subsection{Sampling-Based Approximation to  the NMT Models}
The biggest computational bottleneck of training the NMT models is in the calculation of the softmax layer described in Equation~(\ref{eq: decoder_softmax}), because its computational cost increases linearly with the size of the vocabulary.
The speedup technique with GPUs has proven useful for sequence-based NMT models~\cite{NIPS2014_5346,luong-pham-manning:2015:EMNLP} but it is not easily applicable when dealing with tree-structured data. 
In order to reduce the training cost of the NMT models at the softmax layer, we employ {\it BlackOut}~\cite{DBLP:journals/corr/JiVSAD15}, a sampling-based approximation method.
BlackOut has been shown to be effective in RNN Language Models (RNNLMs) and allows a model to run reasonably fast even with a million word vocabulary with CPUs.

At each word prediction step in the training, BlackOut estimates the conditional probability in Equation~(\ref{eq: output_layer}) for the target word and $K$ negative samples using a weighted softmax function. 
The negative samples are drawn from the unigram distribution raised to the power $\beta\in[0,1]$~\cite{NIPS2013_5021}. 
The unigram distribution is estimated using the training data and $\beta$ is a hyperparameter. 
BlackOut is closely related to Noise Contrastive Estimation (NCE)~\cite{Gutmann:2012:NEU:2503308.2188396} and achieves better perplexity than the original softmax and NCE in RNNLMs. 
The advantages of Blackout over the other methods are discussed in \newcite{DBLP:journals/corr/JiVSAD15}.
Note that BlackOut can be used as the original softmax once the training is finished.

\section{Experiments}
\subsection{Training Data}
We applied the proposed model to the English-to-Japanese translation dataset of the ASPEC corpus given in WAT'15.\footnote{\url{http://orchid.kuee.kyoto-u.ac.jp/WAT/WAT2015/index.html}}
Following \newcite{zhu:2015:WAT}, we extracted the first 1.5 million translation pairs from the training data.
To obtain the phrase structures of the source sentences, i.e., English, we used the probabilistic HPSG parser {\it Enju}~\cite{Miyao:2008:FFM:1350986.1350988}. 
We used Enju only to obtain a binary phrase structure for each sentence and did not use any HPSG specific information.
For the target language, i.e., Japanese, we used KyTea~\cite{neubig-nakata-mori:2011:ACL-HLT2011}, a Japanese segmentation tool, and performed the pre-processing steps recommended in WAT'15.\footnote{\url{http://orchid.kuee.kyoto-u.ac.jp/WAT/WAT2015/baseline/dataPreparationJE.html}}
We then filtered out the translation pairs whose sentence lengths are longer than 50 and whose source sentences are not parsed successfully.
Table~\ref{table: ASPEC} shows the details of the datasets used in our experiments. 
 \begin{table}[t]
  \begin{center}
  \begin{tabular}{l|r|r} 
  				& Sentences 	& Parsed successfully	\\ \hline
	Train			& 1,346,946	& 1,346,946			\\ \hline
	Development    	& 1,790 		& 1,789				\\ \hline	
	Test			& 1,812	 	& 1,811			  
  \end{tabular}
  \caption{Dataset in ASPEC corpus.}	\label{table: ASPEC}
     \end{center}
\end{table}
We carried out two experiments on a small training dataset to investigate the effectiveness of our proposed model and on a large training dataset to compare our proposed methods with the other systems.

The vocabulary consists of words observed in the training data more than or equal to $N$ times.
We set $N = 2$ for the small training dataset and $N=5$ for the large training dataset. 
The out-of-vocabulary words are mapped to the special token  \lq\lq {\it unk}".  
We added another special symbol ``{\it eos}" for both languages and inserted it at the end of all the sentences.
Table~\ref{table: dataset} shows the details of each training dataset and its corresponding vocabulary size. 
 \begin{table}[t]
  \begin{center}
  \begin{tabular}{l|r|r} 
  						& Train (small) 	& Train (large)	\\ \hline
	sentence pairs			& 100,000		& 1,346,946	\\ \hline
	$|V|$ in English     		& 25,478		& 87,796		\\ \hline	
	$|V|$ in Japanese		& 23,532		& 65,680  
  \end{tabular}
  \caption{Training dataset and the vocabulary sizes.}	\label{table: dataset}
     \end{center}
\end{table}

\subsection{Training Details}
The biases, softmax weights, and BlackOut weights are initialized with zeros. 
The hyperparameter $\beta$ of BlackOut is set to 0.4 as recommended by~\newcite{DBLP:journals/corr/JiVSAD15}.
Following \newcite{conf/icml/JozefowiczZS15}, we initialize the forget gate biases of LSTM and Tree-LSTM with 1.0.
The remaining model parameters in the NMT models in our experiments are uniformly initialized in $[-0.1, 0.1]$.
The model parameters are optimized by plain SGD with the mini-batch size of 128. 
The initial learning rate of SGD is 1.0. 
We halve the learning rate when the development loss becomes worse.
Gradient norms are clipped to 3.0 to avoid exploding gradient problems~\cite{DBLP:journals/corr/abs-1211-5063}.

\paragraph{Small Training Dataset}
We conduct experiments with our proposed model and the sequential attentional NMT model with the input-feeding approach. 
Each model has 256-dimensional hidden units and word embeddings.
The number of negative samples $K$ of BlackOut is set to 500 or 2000.

\paragraph{Large Training Dataset}
Our proposed model has 512-dimensional word embeddings and $d$-dimensional hidden units ($d \in \{512, 768, 1024\}$).
$K$ is set to 2500.

Our code\footnote{\url{https://github.com/tempra28/tree2seq}} is implemented in C++ using the Eigen library,\footnote{\url{http://eigen.tuxfamily.org/index.php}} a template library for linear algebra, and we run all of the experiments on multi-core CPUs.\footnote{16 threads on Intel(R) Xeon(R) CPU E5-2667 v3 @ 3.20GHz}
It takes about a week to train a model on the large training dataset with $d = 512$.

\subsection{Decoding process}
We use beam search to decode a target sentence for an input sentence ${\bm x}$ and calculate the sum of the log-likelihoods of the target sentence ${\bm y} = (y_{1}, \cdots ,y_{m})$ as the beam score:
\begin{eqnarray}
score({\bm x}, {\bm y}) = \sum_{j=1}^{m} \log p(y_{j} | {\bm y}_{<{j}}, {\bm x}).
			\label{eq: sentence_score}
\end{eqnarray}
Decoding in the NMT models is a generative process and depends on the target language model given a source sentence.
The score becomes smaller as the target sentence becomes longer, and thus the simple beam search does not work well when decoding a long sentence~\cite{DBLP:journals/corr/ChoMBB14,pougetabadie-EtAl:2014:SSST-8}.
In our preliminary experiments, the beam search with the length normalization in \newcite{DBLP:journals/corr/ChoMBB14} was not effective in English-to-Japanese translation.
The method in \newcite{pougetabadie-EtAl:2014:SSST-8} needs to estimate the conditional probability $p({\bm x} | {\bm y})$ using another NMT model and thus is not suitable for our work.

In this paper, we use statistics on sentence lengths in beam search. 
Assuming that the length of a target sentence correlates with the length of a source sentence, we redefine the score of each candidate as follows:
\begin{eqnarray}
\!\!\!\!\!\!\!\! score({\bm x}, {\bm y}) &\!\!\! = & \!\!\!  L_{{\bm x}, {\bm y}} + \sum_{j=1}^{m} \log p(y_{j} | {\bm y}_{<{j}}, {\bm x}), \\
			\label{eq: revise_sentence_score}
 L_{{\bm x}, {\bm y}} &\!\!\!  = & \!\!\!  \log p(len({\bm y}) | len({\bm x})), 
			\label{eq: penalty_length}
\end{eqnarray}
where $L_{{\bm x}, {\bm y}}$ is the penalty for the conditional probability of the target sentence length $len({\bm y})$ given the source sentence length $len({\bm x})$. 
It allows the model to decode a sentence by considering the length of the target sentence. 
In our experiments, we computed the conditional probability $p(len({\bm y}) | len({\bm x}))$ in advance following the statistics collected in the first one million pairs of the training dataset.
We allow the decoder to generate up to 100 words.

\subsection{Evaluation}
We evaluated the models by two automatic evaluation metrics, RIBES~\cite{Isozaki:2010:AET:1870658.1870750} and BLEU~\cite{Papineni:2002:BMA:1073083.1073135} following WAT'15. 
We used the KyTea-based evaluation script for the translation results.\footnote{\url{http://lotus.kuee.kyoto-u.ac.jp/WAT/evaluation/automatic_evaluation_systems/automaticEvaluationJA.html}}
The RIBES score is a metric based on rank correlation coefficients with word precision, and the BLEU score is based on $n$-gram word precision and a Brevity Penalty (BP) for outputs shorter than the references. 
RIBES is known to have stronger correlation with human judgements than BLEU in translation between English and Japanese as discussed in ~\newcite{Isozaki:2010:AET:1870658.1870750}.

\section{Results and Discussion} 
\subsection{Small Training Dataset} \label{section: Small_Training_Dataset}
 \begin{table*}[t]
 \begin{center}
  \begin{tabular}{l|r|c|c|c|r} 
  	  & \multicolumn{1}{c|}{$K$} & Perplexity & RIBES & BLEU &Time/epoch (min.) \\ \hline \hline
	Proposed model		  & 500	& 19.6	& 71.8	& 20.0	& 55~~~~~~~~~~~~~\\ 
	Proposed model   		  & 2000	& 21.0	& 72.6	& 20.5	& 70~~~~~~~~~~~~~\\ 
	Proposed model (Softmax) & ---	& 17.9	& 73.2	& 21.8	& 180~~~~~~~~~~~~~\\ \hline
	ANMT~\cite{luong-pham-manning:2015:EMNLP}
				 	 	& 500	& 21.6	& 70.7	&18.5	& 45~~~~~~~~~~~~~\\
	+ reverse input 			& 500 	& 22.6	& 69.8	&17.7	& 45~~~~~~~~~~~~~\\
	ANMT~\cite{luong-pham-manning:2015:EMNLP} 
	  	 				& 2000 	& 23.1 	& 71.5	&19.4	& 60~~~~~~~~~~~~~\\
	+ reverse input 			& 2000	& 26.1	& 69.5	&17.5	& 60~~~~~~~~~~~~~\\\hline
  \end{tabular}
    \caption{Evaluation results on the development data using the small training data. The training time per epoch is also shown, and $K$ is the number of negative samples in BlackOut.}	
    \label{table: BLEU_small_data}
   \end{center}
\end{table*}
Table~\ref{table: BLEU_small_data} shows the perplexity, BLEU, RIBES, and the training time on the development data with the Attentional NMT (ANMT) models trained on the small dataset. 
We conducted the experiments with our proposed method using BlackOut and softmax. 
We decoded a translation by our proposed beam search with a beam size of 20.

As shown in Table~\ref{table: BLEU_small_data}, the results of our proposed model with BlackOut improve as the number of negative samples $K$ increases.
Although the result of softmax is better than those of BlackOut ($K=500, 2000$), the training time of softmax per epoch is about three times longer than that of BlackOut even with the small dataset.

As to the results of the ANMT model, reversing the word order in the input sentence decreases the scores in English-to-Japanese translation, which contrasts with the results of other language pairs reported in previous work~\cite{NIPS2014_5346,luong-pham-manning:2015:EMNLP}.
By taking syntactic information into consideration, our proposed model  improves the scores, compared to the sequential attention-based approach.

We found that better perplexity does not always lead to better translation scores with BlackOut as shown in Table~\ref{table: BLEU_small_data}.
One of the possible reasons is that BlackOut distorts the target word distribution by the modified unigram-based negative sampling where frequent words can be treated as the negative samples multiple times at each training step. 

\paragraph{Effects of the proposed beam search}
Table~\ref{table: Beam_small_data} shows the results on the development data of proposed method with BlackOut ($K = 2000$) by the simple beam search and our proposed beam search. 
The beam size is set to 6 or 20 in the simple beam search, and to 20 in our proposed search.
We can see that our proposed search outperforms the simple beam search in both scores.
Unlike RIBES, the BLEU score is sensitive to the beam size and becomes lower as the beam size increases.
We found that the BP had a relatively large impact on the BLEU score in the simple beam search as the beam size increased.
Our search method works better than the simple beam search by keeping long sentences in the candidates with a large beam size. 

\begin{table}[t]
  \begin{center}
   {\small
  \begin{tabular}{l|c|c|c} 
    						& Beam size 	& RIBES		& BLEU~~(BP)			\\ \hline \hline
    \multirow{2}{*}{Simple BS}	& 6	    		& 72.3	 	& 20.0~~(90.1)
    \\ 
     						& 20	 		& 72.3 		& 19.5~~(85.1)		\\ \hline
    Proposed BS				& 20		   	& {\bf 72.6}	& {\bf 20.5}~~(91.7) \\  \hline
  \end{tabular}
  }
  \caption{Effects of the Beam Search (BS) on the development data.}	\label{table: Beam_small_data}
  \end{center}
\end{table}

\paragraph{Effects of the sequential LSTM units}
We also investigated the effects of the sequential LSTMs at the leaf nodes in our proposed tree-based encoder. 
Table~\ref{table: LSTM_tree-based_encoder} shows the result on the development data of our proposed encoder and that of an attentional tree-based encoder without sequential LSTMs  with BlackOut ($K = 2000$).\footnote{For this evaluation, we used the 1,789 sentences that were successfully parsed by Enju because the encoder without sequential LSTMs always requires a parse tree.} 
The results show that our proposed encoder considerably outperforms the encoder without sequential LSTMs, suggesting that the sequential LSTMs at the leaf nodes contribute to the context-aware construction of the phrase representations in the tree.

\begin{table}[t]
    \begin{center}
 {\small
  \begin{tabular}{l|c|c} 
    							& RIBES		& BLEU    \\ \hline \hline
    Without sequential LSTMs		& 69.4 		& 19.5	\\ \hline
    With sequential LSTMs			& {\bf 72.3}	& {\bf 20.0}\\  \hline
  \end{tabular}
  }
  \caption{Effects of the sequential LSTMs in our proposed tree-based encoder on the development data.}	\label{table: LSTM_tree-based_encoder}
    \end{center}
\end{table}

\subsection{Large Training Dataset}
Table~\ref{table: result_large} shows the experimental results of RIBES and BLEU scores achieved by the trained models on the large dataset. 
We decoded the target sentences by our proposed beam search with the beam size of 20.\footnote{We found two sentences which ends without {\it eos} with $d = 512$, and then we decoded it again with the beam size of 1000 following~\newcite{zhu:2015:WAT}.}  
The results of the other systems are the ones reported in \newcite{nakazawa-EtAl:2015:WAT}.

All of our proposed models show similar performance regardless of the value of $d$.
Our ensemble model is composed of the three models with $d =512, 768$, and $1024$, 
and it shows the best RIBES score among all systems.\footnote{Our ensemble model yields a METEOR~\cite{denkowski:lavie:meteor-wmt:2014} score of 53.6 with language option ``-l other". }

As for the time required for training, our implementation needs about one day to perform one epoch on the large training dataset with $d = 512$. 
It would take about 11 days without using the BlackOut sampling.

\paragraph{Comparison with the NMT models} 
The model of \newcite{zhu:2015:WAT} is an ANMT model~\cite{DBLP:journals/corr/BahdanauCB14} with a bi-directional LSTM encoder, and uses 1024-dimensional hidden units and 1000-dimensional word embeddings. 
The model of \newcite{lee-EtAl:2015:WAT} is also an ANMT model with a bi-directional Gated Recurrent Unit (GRU) encoder, and uses 1000-dimensional hidden units and 200-dimensional word embeddings. 
Both models are sequential ANMT models.
Our single proposed model with $d=512$ outperforms the best result of \newcite{zhu:2015:WAT}'s end-to-end NMT model with ensemble and unknown replacement by $+1.19$  RIBES and by $+0.17$ BLEU scores. 
Our ensemble model shows better performance, in both RIBES and BLEU scores, than that of \newcite{zhu:2015:WAT}'s best system which is a hybrid of the ANMT and SMT models by $+1.54$ RIBES and by $+0.74$ BLEU scores and \newcite{lee-EtAl:2015:WAT}'s ANMT system with special character-based decoding by $+1.30$ RIBES and $+1.20$ BLEU scores. 

\paragraph{Comparison with the SMT models}
PB, HPB and T2S are the baseline SMT systems in WAT'15: a phrase-based model, a hierarchical phrase-based model, and a tree-to-string model, respectively~\cite{nakazawa-EtAl:2015:WAT}. 
The best model in WAT'15 is \newcite{neubig-morishita-nakamura:2015:WAT}'s tree-to-string SMT model enhanced with reranking by ANMT using a bi-directional LSTM encoder. 
Our proposed end-to-end NMT model compares favorably with \newcite{neubig-morishita-nakamura:2015:WAT}.

 \begin{table}[t]
 {\small
  \begin{tabular}{l|c|c} 
    Model								& RIBES		& BLEU 		\\ \hline \hline
    Proposed model ($d = 512$)			        & 81.46	 	& 34.36	\\ 
    Proposed model ($d=768$)                   		& 81.89	 	& 34.78	\\
    Proposed model ($d=1024$)             	   	 & 81.58	 	& 34.87	\\
    Ensemble of the above three models				& {\bf 82.45}	& 36.95	\\ \hline \hline
    ANMT with LSTMs~\cite{zhu:2015:WAT}	& 79.70	& 32.19		\\ 
    	+ Ensemble, {\it unk} replacement 		& 80.27	& 34.19		\\
        + System combination,
        									& \multirow{2}{*}{80.91} & \multirow{2}{*}{36.21}  \\ 
	~~~ 3 pre-reordered ensembles		&	& \\ \hline
    ANMT with GRUs~\cite{lee-EtAl:2015:WAT}
    									& \multirow{3}{*}{81.15} & \multirow{3}{*}{35.75}		\\ 
    	+ character-based decoding,  		& 	& 	\\ 
	~~~ Begin/Inside representation		& 	& \\ \hline \hline
    PB baseline							& 69.19	& 29.80		\\
    HPB baseline						& 74.70	& 32.56		\\
    T2S baseline							& 75.80	& 33.44		\\  \hline
    T2S model~\cite{neubig-duh:2014:P14-2}
    									& 79.65	& 36.58		\\
     + ANMT Rerank~\cite{neubig-morishita-nakamura:2015:WAT}
    									& 81.38	& 38.17		\\ \hline 			

  \end{tabular}
  }
  \caption{Evaluation results on the test data.}	\label{table: result_large}
\end{table}

\subsection{Qualitative Analysis}
\begin{figure}[t]
  \begin{center}
  \includegraphics[clip,width=7.8cm]{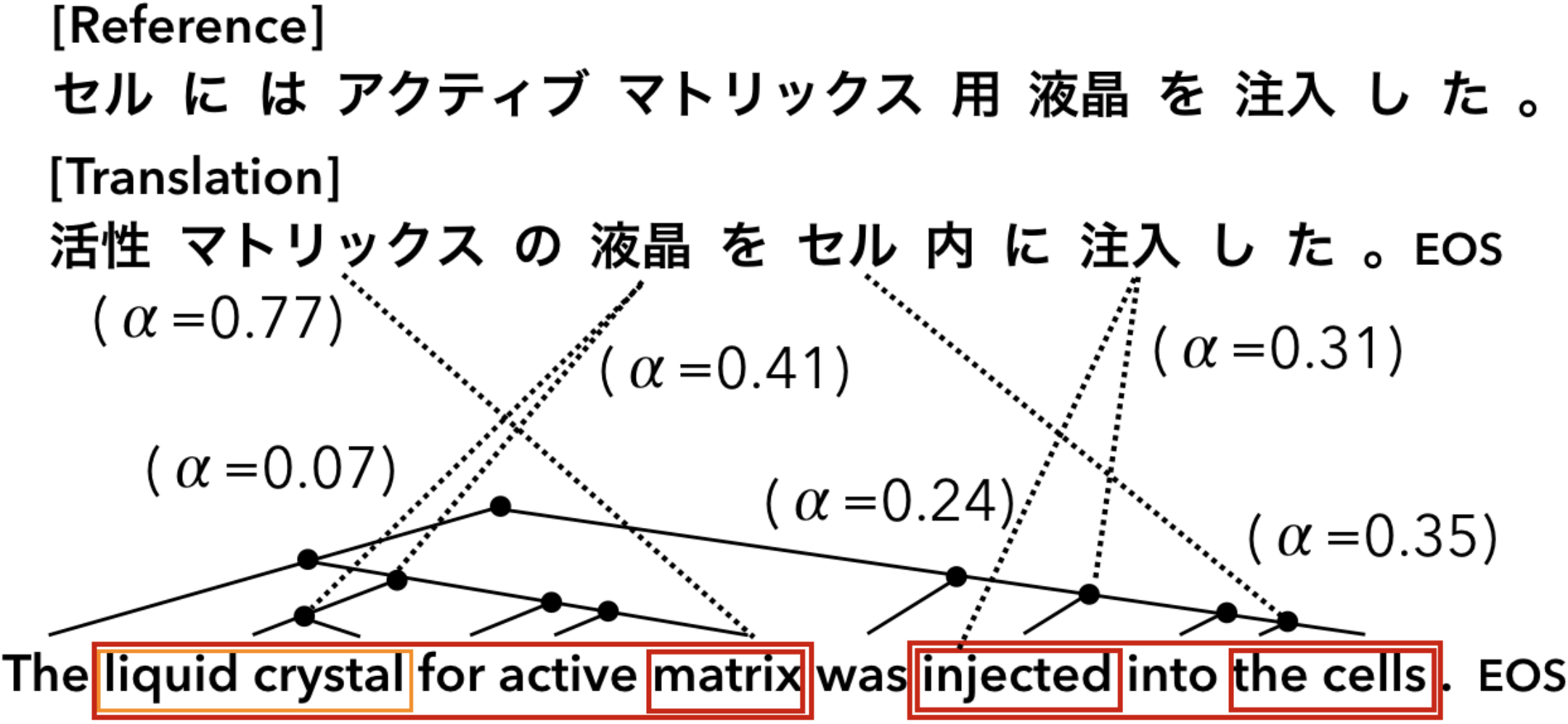}
    \caption{Translation example of a short sentence and the attentional relations by our proposed model.}
   \label{fig: example_translation}
  \end{center}
\end{figure}

\begin{figure*}[ht]
  \begin{center}
  \includegraphics[clip,width=16cm]{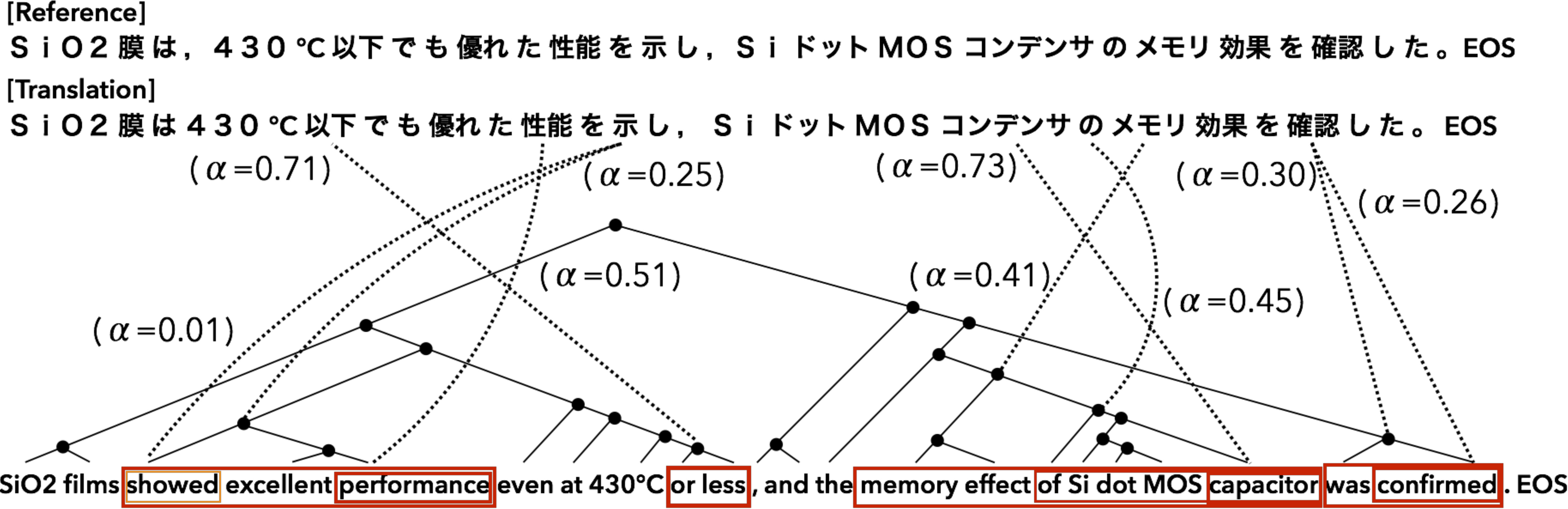}
    \caption{Translation example of a long sentence and the attentional relations by our proposed model.}
   \label{fig: example_translation2}
  \end{center}
\end{figure*}

We illustrate the translations of test data by our model with $d = 512$ and several attentional relations when decoding a sentence. 
In Figures~\ref{fig: example_translation} and \ref{fig: example_translation2}, an English sentence represented as a binary tree is translated into Japanese, and several attentional relations between English words or phrases and Japanese word are shown with the highest attention score $\alpha$. 
The additional attentional relations are also illustrated for comparison.
We can see the target words softly aligned with source words and phrases.

In Figure~\ref{fig: example_translation}, the Japanese word \begin{CJK}{UTF8}{min}``液晶"\end{CJK} means ``liquid crystal", and it has a high attention score ($\alpha=0.41$) with the English phrase ``liquid crystal for active matrix".
This is because the $j$-th target hidden unit ${\bm s}_{j}$ has the contextual information about the previous words ${\bm y_{<j}}$ including \begin{CJK}{UTF8}{min}``活性 マトリックス の"\end{CJK} (``for active matrix" in English).
The Japanese word \begin{CJK}{UTF8}{min}``セル"\end{CJK} is softly aligned with the phrase ``the cells" with the highest attention score ($\alpha=0.35$).
In Japanese, there is no definite article like ``the" in English, and it is usually aligned with {\it null} described as Section~\ref{Introduction}. 

In Figure~\ref{fig: example_translation2}, in the case of the Japanese word \begin{CJK}{UTF8}{min}``示"\end{CJK} (``showed" in English), the attention score with the English phrase ``showed excellent performance" ($\alpha=0.25$) is higher than that with the English word ``showed" ($\alpha=0.01$).
The Japanese word \begin{CJK}{UTF8}{min}``の"\end{CJK} (``of" in English) is softly aligned with the phrase ``of Si dot MOS capacitor" with the highest attention score ($\alpha=0.30$).
It is because our attention mechanism takes each previous context of the Japanese phrases \begin{CJK}{UTF8}{min}``優れ た 性能"\end{CJK} (``excellent performance" in English) and \begin{CJK}{UTF8}{min}``Ｓｉ ドット ＭＯＳ コンデンサ"\end{CJK} (``Si dot MOS capacitor" in English) into account and softly aligned the target words with the whole phrase when translating the English verb ``showed" and the preposition ``of".
Our proposed model can thus flexibly learn the attentional relations between English and Japanese.

We observed that our model translated the word ``active" into \begin{CJK}{UTF8}{min}``活性"\end{CJK}, a synonym of the reference word \begin{CJK}{UTF8}{min}``アクティブ"\end{CJK}. 
We also found similar examples in other sentences, where our model outputs synonyms of the reference words, e.g. \begin{CJK}{UTF8}{min}``女"\end{CJK} and \begin{CJK}{UTF8}{min}``女性"\end{CJK} (``female" in English) and ``NASA" and \begin{CJK}{UTF8}{min}``航空宇宙局"\end{CJK} (``National Aeronautics and Space Administration" in English).
These translations are penalized in terms of BLEU scores, but they do not necessarily mean that the translations were wrong. 
This point may be supported by the fact that the NMT models were highly evaluated in WAT'15 by crowd sourcing~\cite{nakazawa-EtAl:2015:WAT}. 

\section{Related Work}
\newcite{kalchbrenner-blunsom:2013:EMNLP} were the first to propose an end-to-end NMT model using Convolutional Neural Networks (CNNs) as the source encoder and using RNNs as the target decoder.
The Encoder-Decoder model can be seen as an extension of their model, and it replaces the CNNs with RNNs using GRUs~\cite{cho-EtAl:2014:EMNLP2014} or LSTMs~\cite{NIPS2014_5346}.

\newcite{NIPS2014_5346} have shown that making the input sequences reversed is effective in a French-to-English translation task, and the technique has also proven effective in translation tasks between other European language pairs~\cite{luong-pham-manning:2015:EMNLP}.
All of the NMT models mentioned above are based on sequential encoders. 
To incorporate structural information into the NMT models, \newcite{DBLP:journals/corr/ChoMBB14} proposed to jointly learn structures inherent in source-side languages but did not report improvement of translation performance.
These studies motivated us to investigate the role of syntactic structures explicitly given by existing syntactic parsers in the NMT models.

The attention mechanism~\cite{DBLP:journals/corr/BahdanauCB14} has promoted NMT onto the next stage.
It enables the NMT models to translate while aligning the target with the source.
\newcite{luong-pham-manning:2015:EMNLP} refined the attention model so that it can dynamically focus on local windows rather than the entire sentence. 
They also proposed a more effective attentional path in the calculation of ANMT models. 
Subsequently, several ANMT models have been proposed~\cite{DBLP:journals/corr/ChengSHHWSL15,corr/1601.01085}; however, each model is based on the existing sequential attentional models and does not focus on a syntactic structure of languages.

\section{Conclusion}
In this paper, we propose a novel syntactic approach that extends attentional NMT models.
We focus on the phrase structure of the input sentence and build a tree-based encoder following the parsed tree.
Our proposed tree-based encoder is a natural extension of the sequential encoder model, where the leaf units of the tree-LSTM in the encoder can work together with the original sequential LSTM encoder.
Moreover, the attention mechanism allows the tree-based encoder to align not only the input words but also input phrases with the output words.
Experimental results on the WAT'15 English-to-Japanese translation dataset demonstrate
that our proposed model achieves the best RIBES score and outperforms the sequential attentional NMT model.

\section*{Acknowledgments}
We thank the anonymous reviewers for their constructive comments and suggestions.
This work was supported by CREST, JST, and JSPS KAKENHI Grant Number 15J12597.
\bibliography{acl2016}
\bibliographystyle{acl2016}

\appendix
\end{document}